\UseRawInputEncoding
\pdfoutput=1
\pdfoutput=1
\pdfoutput=1
\pdfoutput=1
\documentclass[sigconf]{acmart}

\copyrightyear{2023} 
\acmYear{2023} 
\setcopyright{acmlicensed}\acmConference[KDD '23]{Proceedings of the 29th ACM SIGKDD Conference on Knowledge Discovery and Data Mining}{August 6--10, 2023}{Long Beach, CA, USA}
\acmBooktitle{Proceedings of the 29th ACM SIGKDD Conference on Knowledge Discovery and Data Mining (KDD '23), August 6--10, 2023, Long Beach, CA, USA}
\acmPrice{15.00}
\acmDOI{10.1145/3580305.3599331}
\acmISBN{979-8-4007-0103-0/23/08}
\begin{document}

\title{Empowering General-purpose User Representation with Full-life Cycle Behavior Modeling}

\author{Bei Yang}

\authornote{Both authors contributed equally to this research.}
\email{bella.yb@alibaba-inc.com}
\orcid{0000-0002-1997-234X}
\affiliation{%
  \institution{Alibaba Group}
  \city{Hangzhou}
  \country{China}
  \postcode{310000}
}

\author{Jie Gu}
\authornotemark[1]
\orcid{0000-0002-9753-269X}
\email{yemu.gj@alibaba-inc.com}
\affiliation{%
  \institution{Alibaba Group}
  \city{Hangzhou}
  \country{China}
  \postcode{310000}
}

\author{Ke Liu}
\email{ke.l@wustl.edu}
\affiliation{%
  \institution{Zhejing University}
  \city{Hangzhou}
  \country{China}
}

\author{Xiaoxiao Xu}
\email{xiaoxiao.xuxx@alibaba-inc.com}
\affiliation{%
 \institution{Alibaba Group}
 \city{Hangzhou}
 \country{China}
 }

\author{Renjun Xu}
\email{rux@zju.edu.cn}
\affiliation{%
  \institution{Zhejing University}
  \city{Hangzhou}
  \country{China}
  }

\author{Qinghui Sun}
\email{yuyang.sqh@alibaba-inc.com}
\affiliation{%
 \institution{Alibaba Group}
 \city{Hangzhou}
 \country{China}
 }

\author{Hong Liu}
\email{229795716@qq.com}
\affiliation{%
 \institution{Alibaba Group}
 \city{Hangzhou}
 \country{China}
 }

\renewcommand{\shortauthors}{Bei Yang, et al.}

\begin{abstract}
  User Modeling plays an essential role in industry. In this field, task-agnostic approaches, which generate general-purpose representation applicable to diverse downstream user cognition tasks, is a promising direction being more valuable and economical than task-specific representation learning. 
  With the rapid development of Internet service platforms, user behaviors have been accumulated continuously. However, existing general-purpose user representation researches have little ability for full-life cycle modeling on extremely long behavior sequences since user registration.
  In this study, we propose a novel framework called full- Life cycle User Representation Model (LURM) to tackle this challenge. Specifically, LURM consists of two cascaded sub-models: (\romannumeral1) Bag-of-Interests (BoI) encodes user behaviors in any time period into a sparse vector with super-high dimension (\textit{e}.\textit{g}., $10^{5}$); (\romannumeral2) Self-supervised Multi-anchor Encoder Network (SMEN) maps sequences of BoI features to multiple low-dimensional user representations. Specially, SMEN achieves almost lossless dimensionality reduction, benefiting from a novel multi-anchor module which can learn different aspects of user interests. Experiments on several benchmark datasets show that our approach outperforms state-of-the-art general-purpose representation methods.
\end{abstract}




\maketitle

\section{Introduction}
Customer first is a well-known value insisted by many businesses. Accordingly, understanding users is of great importance since it helps provide satisfactory personalized services. Researchers have reached a common view that user historical behaviors contain rich valuable information\cite{dupret2008user,he2014practical,elkahky2015multi,yu2016dynamic}. It is intuitive to mine and capture user interests or properties from massive behavior data. A typical solution is to encode behavior sequences into low-dimensional yet informative representations. Extensive works have proved the success of such a solution, benefiting a wide range of real-world applications like user profiling, recommendation systems, search engines and online advertising.

In the literature, there are plenty of works focusing on task-specific user representation learning. That is, the representation is simultaneously learned with the specific downstream classifier, \emph{e.g.}, CTR predictor for recommendation \cite{ren2019lifelong,yu2019adaptive,ji2020two}. Such a task-specific paradigm would benefit the performance on the target application, but the learned representations can hardly be applied to other tasks. To avoid generalization issues and ensure performance, training a particular model for each downstream application is essential. However, it is time-consuming, expensive (requiring massive labeled data, computing and storage resources), and cannot be used in many real-world scenarios where business requirements are diverse, variable and numerous.

In contrast, general-purpose (\emph{a.k.a.}, universal) user representation has gained growing attention recently\cite{robertson2004understanding,ding2017multi,ni2018perceive,andrews2019learning,gu2020exploiting,wu2020ptum,2021Interest,2021Scaling,cao2022sampling,wu2022userbert}. 
Typically, universal user representation is learned without any task-specific bias and can serve a variety of downstream tasks by treating the pre-trained model as a feature extractor. 
It can be extracted from user historical behavior data or other side information to express his/her interests and preferences. Note that no more rectifications (e.g., fine-tuning) are required for the representation model in downstream applications. We only need to train a simple model, like SVM or MLP, for a specific task. Unfortunately, current approaches can only process user behavior sequences with a length of tens or hundreds. According to our observations, the performance of the universal representation model is limited by the available behaviors.

An insight we want to show is that the more behaviors, the richer the information can be captured, and the better the performance on downstream (Fig. \ref{comparison_length}). Several previous works have made some attempts on this topic \cite{kumar2016ask,dai2019transformer,yang2019xlnet}. They model relatively long sequential data based on hierarchical architectures and memory networks \cite{ying2018sequential,pi2019practice,ren2019lifelong}, and verify the effectiveness of incorporating more behaviors. However, this topic is far from being solved. The existing methods suffer from the generalization issue (since they are all task-specific), as well as the lack of the capability of handling extremely long behavior sequences. With the rapid development of Internet service platforms, abundant user behavior sequences which reflect intrinsic and multi-facet user interests have been accumulated. We do believe that valuable and rich information can be mined from such massive data. The core is to efficiently encode full-life cycle behaviors of users (which may even include hundreds of thousands of behaviors) into informative general-purpose representations. 



In this work, a novel framework for universal user modeling called full-Life cycle User Representation Model (LURM) is proposed. The framework has the ability to model user behaviors of arbitrary length since his/her registration, e.g., including every behavior from his/her registration on some APP to the present day. To meet the need of extremely long sequence modeling, we first introduce a model named Bag-of-Interests (BoI) to summarize items in behavior sequences similar to Bag of Visual Words. In this way, we can use a sparse vector with super-high dimension to represent user behaviors in any time period. Then, a Self-supervised Multi-anchor Encoder Network (SMEN) that maps sequences of BoI features to multiple low-dimensional user representations is proposed. SMEN consists of three modules: a multi-anchor module which can learn different aspects of user preferences, a time aggregation module which can model evolution of user behaviors, and a multi-scale aggregation module which can learn and aggregate BoI features in different scales. Considering the consistency between user behaviors in different time periods, we introduce a contrastive loss function to the self-supervised training of SMEN. 
With the designs above, SMEN achieves almost lossless dimensionality reduction. 
It is noteworthy that the proposed method allows for encoding behaviors of arbitrary length and aggregating information in different time scales. Thus, though the inspiration of this work is to capture long-term user interests from full-life cycle behaviors, LURM can also be applied to short-term interests related tasks. Extensive experiments on several benchmark datasets show the superiority of our approach against other baselines on both short-term interest-related tasks and long-term interest-related tasks.

\begin{figure*}[ht]
\begin{center}
\centerline{\includegraphics[width=0.9\textwidth]{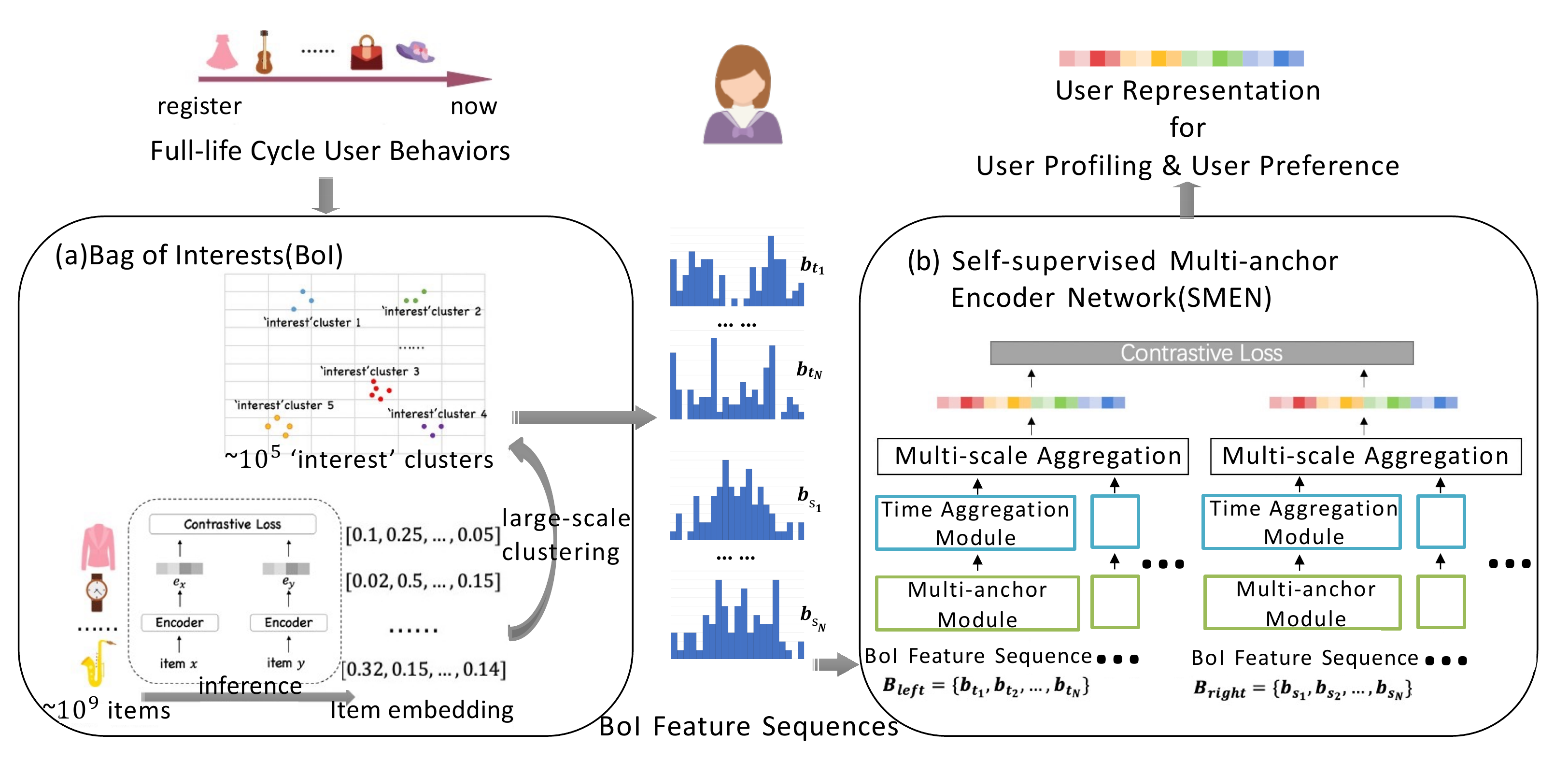}}
\caption{Illustration of our full-Life cycle User Representation Model (LURM) for user understanding. LURM consists of two sub-model: (a) Bag-of-Interests (BoI) is used to aggregate user behavior data at `interest' granularity, and is composed of an item embedding module and a large scale clustering module. We apply the BoI model on full-life cycle behavior data to generate multi-scale BoI feature sequences, (b) Self-supervised Multi-anchor Encoder Network (SMEN) is used to learn compressed user representations from BoI features sequences, and is composed of a multi-anchor module, a time aggregation module, a multi-scale aggregation module and a contrastive learning module.}
\label{LURMoverflow}
\end{center}
\end{figure*}

The main contribution of our work can be summarized as follows: 
\begin{itemize}
\item[$\bullet$] A novel framework named LURM is proposed for learning high-quality universal user representations via self-supervision. The framework is built based on Bag-of-Interests, which is capable of encoding arbitrary numbers of user behaviors. It shows great advantages in adaptively modeling long-term or relatively short-term user interests in a data-driven manner.

\item[$\bullet$] More importantly, such a Bag-of-Interests formulation allows us to encode extremely long behavior sequences (even including millions of behaviors). This makes full-life cycle user modeling a reality, while without forgetting and efficiency issues as in previous RNN based or Transformer based methods.

\item[$\bullet$] A sub-module named SMEN is proposed, which further incorporates interest-anchored dimension reduction and time variation modeling. Such operations ensure a compact, easy-using and informative universal user representation.

\item[$\bullet$] Extensive experiments are performed on several real-world datasets. The results demonstrate the effectiveness and generalization ability of the learned user representation. Furthermore, the experiments also demonstrate that better performance can indeed be achieved by full-life cycle behavior modeling.
\end{itemize}

\section{Methodology}
In this work, we are committed to learning general-purpose user representation by modeling full-life cycle user behavior sequences with arbitrary length. For this purpose, we propose a framework named full-Life cycle User Representation Model (LURM), which consists of two cascaded sub-models, \emph{i.e.}, Bag-of-Interests (BoI) and Self-supervised Multi-anchor Encoder Network (SMEN). The overall architecture of LURM is shown in Fig. \ref{LURMoverflow}.

\subsection{Bag-of-Interests}
Most of the time, there are certain patterns behind user behaviors, which are supposed to be focused on to mine user interests and preferences. To this end, we should encode behaviors (like purchase and click) by aggregating the contents of items. Bag of Words (BoW) \cite{harris1954distributional} is a common method, which aggregates information of items at the word granularity, \emph(e.g.), with item titles. However, it cannot deal with the data of other modalities such as images, nor can it model the dependencies between words. Inspired by Bag of Visual Words (BoVW) \cite{fei2005bayesian}, encoding behaviors at the item granularity seems more natural and reasonable. Unfortunately, it is also inappropriate in the field of e-commerce, since usually there are billions of items and accordingly the item vocabulary would be extremely large, making it infeasible in practice. 

We propose a model called Bag-of-Interests (BoI) to aggregate users’ behavior data at the `interest' granularity. Each `interest' is a cluster of similar items and represents a certain kind of preference. The size of the `interest' vocabulary is often selected at a level of about $10^5$ for retaining enough details. As shown in Fig. \ref{LURMoverflow} (a), BoI consists of an item embedding module and a large-scale clustering module. For convenience, we only focus on the text modality in this work. It should be noted that our method can be easily extended to multi-modal data.

\subsubsection{Item Embedding Module}
\label{contrastive loss}
An `interest' vocabulary is supposed to be built in our BoI model, similar to BoVW. The embedding of each item is required, so that similar items with close distance in the embedding space can be clustered together to form an `interest'. Recently, discriminative approaches based on contrastive learning in the latent space have shown great success in the field of representation learning, achieving state-of-the-art results in natural language and image processing\cite{mikolov2013distributed,devlin2018bert,oord2018representation,chen2020simple}. 
Inspired by these works, we design a contrastive learning task based on the relation between items drawn from a user to learn item embedding\cite{barkan2016item2vec}.

Given a set of users $U=\{u_1, u_2,...,u_{|U|}\}$, each user $u\in U$ corresponds to a behavior sequence $\boldsymbol{S}=\left\{\boldsymbol{x_1}, \boldsymbol{x_2},..., \boldsymbol{x_{|S|}}\right\}$, where $\boldsymbol{x_i} \in \boldsymbol{S}$ denotes the $i$-th item. $|\boldsymbol{U}|$ and $|\boldsymbol{S}|$ denote the number of users and the length of $u$'s behaviors respectively. Generally, the content of an item $\boldsymbol{x}$ can be expressed as $\{w_{1},w_{2},...,w_{|\boldsymbol{x}|}\}$, where $w_i$ denotes a word from a vocabulary $V$, and $|\boldsymbol{x}|$ denotes the number of words in the content of $\boldsymbol{x}$. Firstly, an encoder with average operation is used to generate item embedding $e$:
\begin{equation}
\label{ItemEncoder}
\boldsymbol{e_x}=encoder(w_{1},w_{2},...,w_{|\boldsymbol{x}|})=proj(\frac{1}{|\boldsymbol{x}|}\sum_{i=1}^{|\boldsymbol{x}|}\boldsymbol{W_i}),
\end{equation}
where $\boldsymbol{W_i}\in \mathbb{R}^d$ is the embedding of word $w_i$ and will be learned during training, $proj(\cdot)$ includes two residual blocks, and a $L_2$ normalization layer. To construct the contrastive learning task, we sample positive pairs from behavior sequences of users randomly. Specifically, two items $(\boldsymbol{x_i},\boldsymbol {y_i})$ are similar, i.e. a positive pair, if they are drawn from the same user behavior sequence and the time interval between the occurrence of these two items is less than $\beta$, where $\beta$ is the window size controlling the interval of the two user behaviors. Without loss of generality, the sampled mini-batch with batch size $n$ can be denoted as $\boldsymbol{\Delta}=\{\boldsymbol{x^1},\boldsymbol{y^1},\boldsymbol{x^2},\boldsymbol{y^2},...,\boldsymbol{x^n},\boldsymbol{y^n}\}$, where $(\boldsymbol{x^i},\boldsymbol{y^i})$ construct a positive pair drawn from the behavior sequence $\boldsymbol{S^{i}_t}$ of the $i$-th user in batch. Then, the contrastive prediction task is defined to identify $\boldsymbol{y^i}$ in $\boldsymbol{\Delta}\setminus \{\boldsymbol{x^i}\}$ for a given $\boldsymbol{x^i}$, and all other items in $\boldsymbol{\Delta}\setminus \{\boldsymbol{x^i},\boldsymbol{y^i}\}$ are negatives. The loss for the positive pair $(\boldsymbol{x^i},\boldsymbol{y^i})$ is written as
\begin{equation}
\label{positive loss}
l(\boldsymbol{x^i},\boldsymbol{y^i})=-\log \frac{e^{g(\boldsymbol{x^i},\boldsymbol{y^i})/\tau}}{\sum_{\boldsymbol{\nu}\in \boldsymbol{\Delta}, \boldsymbol{\nu}\neq \boldsymbol{x^i}}e^{g(\boldsymbol{x^i},\boldsymbol{\nu})/\tau}},
\end{equation}
where $g(\boldsymbol{x},\boldsymbol{y})=\frac{\boldsymbol{e_{x}}^T\boldsymbol{e_{y}}}{\|\boldsymbol{e_x}\|\|\boldsymbol{e_y}\|}=\boldsymbol{e_{x}}^T\boldsymbol{e_{y}}$ denotes the cosine similarity between the embedding $\boldsymbol{e_{x}}$ and the embedding $\boldsymbol{e_{y}}$ , and $\tau$ is the temperature parameter. The final objective is the average loss of all positive pairs in the mini-batch, which can be written as
\begin{equation}
\label{contrastiveloss}
Loss=\frac{1}{2n}\sum_i(l(\boldsymbol{x^i},\boldsymbol{y^i})+l(\boldsymbol{y^i},\boldsymbol{x^i})).
\end{equation}


\subsubsection{Large-scale Clustering Module}
An item embedding set $\boldsymbol{E}=\left\{\boldsymbol{e_i}\right\}_{i\in I}$ can be obtained, where $I$ is the complete collection of items at the billion level. In order to retain details as many as possible, the size of the `interest' vocabulary is set to be at $10^4\sim 10^5$ level. In other words, all items should be clustered into $D$ (e.g., $10^5$) categories. Considering the large-scale item set, a subset $\boldsymbol{E'} \subset \boldsymbol{E}$ at the million level is sampled, and an efficient clustering algorithm named HDSC \cite{yi2014single} on $\boldsymbol{E'}$ is employed to cluster similar items into the same `interest'. 

After clustering, the cluster centers $\boldsymbol C$ make up an `interest' vocabulary. Therefore, each item can be attributed to one/multiple `interest(s)' by hard/soft cluster assignment. Take hard cluster assignment as an example, each user can obtain his/her sparsely high-dimensional BoI feature $\boldsymbol{b_t}\in \mathbb R ^D$ in time period $\boldsymbol{t}$ according to his/her behavior sequence $\boldsymbol{S_t}=\left\{\boldsymbol{x_1}, \boldsymbol{x_2},..., \boldsymbol{x_{|\boldsymbol{S_t}|}}\right\}$ :
\begin{equation}
\label{BoI}
\begin{split}
\boldsymbol{b_t}=[log(1+\Sigma_{i=1}^{|\boldsymbol{S_t}|} \mathbb I_{\boldsymbol{x_i}\in c_1}), log(1+\Sigma_{i=1}^{|\boldsymbol{S_t}|} \mathbb I_{\boldsymbol{x_i}\in c_2}),\\ ..., log(1+\Sigma_{i=1}^{|\boldsymbol{S_t}|} \mathbb I_{\boldsymbol{x_i}\in c_D})]
\end{split}
\end{equation}
where $\boldsymbol{x_i}\in c_j$ means item $i$ is assigned to the cluster $c_j\in \boldsymbol C$, $\mathbb I$ is an indicator function.

\begin{figure*}[ht]
\begin{center}
\centerline{\includegraphics[width=0.9\textwidth]{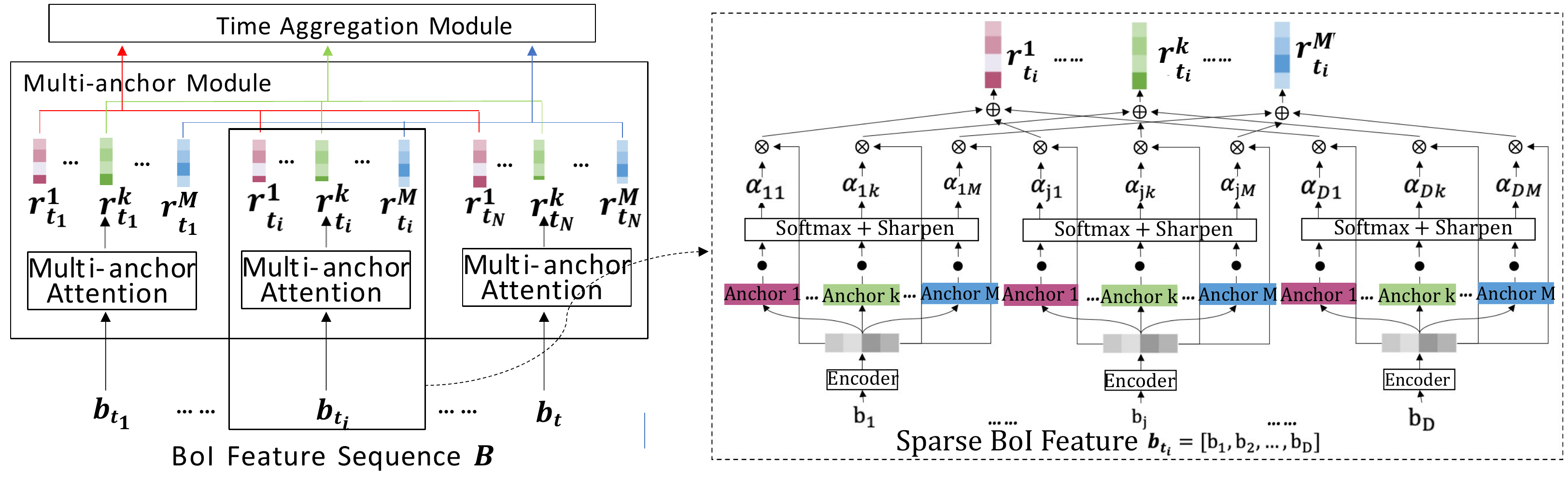}}
\caption{Illustration of the multi-anchor module in SMEN. The module outputs a group of diverse representations by assigning different portions of each `interest' to different anchors. The $\bullet$ denotes the dot product operation, $\otimes$ denotes the scalar multiply operation, and $\oplus$ denotes vector sum operation.
}
\label{multi anchor}
\end{center}
\end{figure*}

\subsection{Self-supervised Multi-anchor Encoder Network}
A BoI feature $\boldsymbol{b_t}$ for each user can be obtained, through the BoI model given behavior data in any time period $\boldsymbol t$. Most directly, a user representation with super-high dimension $\boldsymbol{b_T}$ can be obtained by applying the BoI model to the whole life-cycle time $\boldsymbol T$. However, there are two main disadvantages, \emph{i.e.}, 1)  it is not friendly to the downstream tasks since the dimension of representation is too high, and 2) it is too crude to aggregate the information in the whole life-cycle time without considering variations over time.

Therefore, we propose to get a BoI feature sequence by applying the BoI model at each time period of the whole life-cycle time. Then, a Self-supervised Multi-anchor Encoder Network (SMEN) is designed to learn compressed user representations from the sequence of BoI features. {According to its design, SMEN can simultaneously generate multiple low-dimensional representations (e.g., each user representation has dimension of $10^2$) with a certain degree of disentanglement, representing different aspects of user preferences.}

The full-life cycle time $\boldsymbol T$ is divided into $N$ parts at a fixed time interval, i.e. $\boldsymbol T=\left\{\boldsymbol{t_1}, \boldsymbol{t_2}, ..., \boldsymbol{t_N}\right\}$ and $\boldsymbol{t_i} $ denotes the $i$-th time period. In our experiments, the time interval is usually set to be monthly/seasonly/yearly granularity. In this way, a sequence of BoI features $\boldsymbol B=\left\{\boldsymbol{b_{t_1}}, \boldsymbol{b_{t_2}}, ..., \boldsymbol{b_{t_N}}\right\}$ can be obtained, where $\boldsymbol{b_{t_i}}$ denotes the BoI feature corresponding to the $i$-th time period $\boldsymbol{t_i}$. After that, SMEN is used to map $\boldsymbol B$ to low-dimensional user representations. As shown in Fig. \ref{LURMoverflow} (b), SMEN consists of a multi-anchor module, a time aggregation module, a multi-scale aggregation module and a contrastive learning module. Details of the model will be described in the following subsections.

\subsubsection{Multi-anchor Module}
The data of user behavior, which is highly unstructured and complex, implies different preferences of the user. To capture diverse aspects of user preferences, a novel multi-anchor module is proposed. {Due to the fact that each user has different interests, and the degree of preference for different interests is also different, the design of multi-anchor module is quite different from multi-head attention.} Specifically, suppose there are $M$ anchors, and each of them indicates a certain preference of users. Let $\boldsymbol{b}$ be a BoI feature, the module converts $\boldsymbol{b}$ to $M$ low-dimensional representations as shown in Fig. \ref{multi anchor}. Each representation is computed as
\begin{equation}
\label{MultiAnchor}
\boldsymbol{r^i}=ReLU(\boldsymbol{\alpha_i}^Tf(\boldsymbol{b}))=ReLU(\sum_j^D\alpha_{ij}f(b_j))=ReLU(\sum_j^D\alpha_{ij}b_j \boldsymbol{W^e_j}),
\end{equation}
where $f(b_j)=b_j\boldsymbol{W^e_j}$ is the `interest' embedding function, $b_j$ is the $j$-th element of $\boldsymbol{b}$, $\boldsymbol{W^e}={(\boldsymbol{W^e_1}, \boldsymbol{W^e_2}, ..., \boldsymbol{W^e_D})}^T$ is the embedding matrix. And $\alpha_{ij}$ is the attention weight between the $i$-th anchor and the $j$-th `interest', which measures the portion assigned to the $i$-th preference from the $j$-th behavior `interest'. The weight $\alpha_{ij}$ is defined as
\begin{equation}
\label{Attention}
\alpha_{ij} = \frac{\exp(\boldsymbol{W^a_i k_j})}{\sum_l\exp(\boldsymbol{W^a_l k_j})},
\end{equation}
where $\boldsymbol{W^a_i}$ is the vector corresponding to the $i$-th anchor, $\boldsymbol{W^a}={(\boldsymbol{W^a_1}, \boldsymbol{W^a_2}, ..., \boldsymbol{W^a_M})}^T$ is the anchor matrix, and $\boldsymbol{k_j}=\boldsymbol{W^p}ReLU(\boldsymbol{W_j^e})$ is the interest vector corresponding to the $j$-th `interest'. $\boldsymbol{W^e}\in \mathbb{R}^{D\times H}$, $\boldsymbol{W^a}\in \mathbb{R}^{M\times H}$, and $\boldsymbol{W^p}\in \mathbb{R}^{H\times H}$ are learned parameters. $\boldsymbol{r^i}$ can be computed efficiently since $\boldsymbol{b}$ is a sparse vector. Due to the different anchor vectors, different attention weights can be generated for each `interest'. Finally, a group of different aggregated representations $\boldsymbol{R}=\{\boldsymbol{r^1}, \boldsymbol{r^2}, ..., \boldsymbol{r^M}\}$ can be obtained ($ \boldsymbol{M}$ indicates the total number of anchors). In this way, we can learn different aspects of user preferences. 
Specially, experiments prove that SMEN can achieve almost lossless dimensionality reduction mainly due to the multi-anchor module.

\subsubsection{Time Aggregation Module}
Through the multi-anchor module, a sequence of representation groups $\boldsymbol{\mathcal{R}}=\{\boldsymbol{R_{t_1}}, \boldsymbol{R_{t_2}}, ..., \boldsymbol{R_{t_N}}\}$ for each user is obtained, where $\boldsymbol{R_{t_i}}=\{\boldsymbol{r_{t_i}^1}, \boldsymbol{r_{t_i}^2}, ..., \boldsymbol{r_{t_i}^M}\}$ is  a representation group generated by multi-anchor module corresponding to the BoI feature $\boldsymbol{b_{t_i}}$ in time period $\boldsymbol{t_i}$. In the time aggregation module, each sequence $\boldsymbol{R^i}=\{\boldsymbol{r_{t_1}^i}, \boldsymbol{r_{t_2}^i}, ..., \boldsymbol{r_{t_N}^i}\}$ is aggregated separately to yield a new representation $\boldsymbol{\tilde{r}^i}\in \mathbb R^H$. Thus the variable-size representation sequence $\boldsymbol{\mathcal{R}}$ can be transformed into $M$ fixed-size representations $\boldsymbol{\tilde R}=\left\{\boldsymbol{\tilde{r}^1}, \boldsymbol{\tilde{r}^2}, ..., \boldsymbol{\tilde{r}^M}\right\}$.

There are various methods which can be used to aggregate variable-size sequences, such as average/max pooling and RNNs. Compared to average and max pooling, RNNs are more appropriate to capture variations over time.
Among all RNN-based models, long short-term memory (LSTM) and gated recurrent units (GRU) are most commonly used. Considering that GRU has fewer parameters and is less computationally intensive, GRU is adopted to the time aggregation module in this work. 


\subsubsection{Multi-scale Aggregation Module}
Under a time division $\boldsymbol T=\left\{\boldsymbol{t_1}, \boldsymbol{t_2}, ..., \boldsymbol{t_N}\right\}$, 
we have obtained $M$ representations $\boldsymbol{\tilde R}= \{\boldsymbol{\tilde{r}^1}, \boldsymbol{\tilde{r}^2}, ...,$ 
$ \boldsymbol{\tilde{r}^M}\}$ 
through the BoI, multi-anchor module and time aggregation module as introduced above. It's worth noting that if the time interval is too small, the input sequence of SMEN, i.e. $\boldsymbol B=\left\{\boldsymbol{b_{t_1}}, \boldsymbol{b_{t_2}}, ..., \boldsymbol{b_{t_N}}\right\}$ will become extremely long, which contains extensive details but causes modeling difficulties due to catastrophic forgetting. On the other hand, details may be lost if the time interval is too large.

To address this trade-off, multi-scale aggregation module is designed. The module captures diverse patterns from user behavior by aggregating several representations generated at different granularities. 
{For example, if user behavior is aggregated at two granularities,} 
i.e. $\boldsymbol T=\left\{\boldsymbol{t_1}, \boldsymbol{t_2}, ..., \boldsymbol{t_N}\right\}=\left\{\boldsymbol{t'_1}, \boldsymbol{t'_2}, ..., \boldsymbol{t'_{N'}}\right\}$ (e.g., if the length of time period is 5 years, $\boldsymbol{N}$ is 60 for monthly granularity, and $\boldsymbol{N'}$ is 5 for yearly granularity). Thus, two sequences of BoI feature $\boldsymbol B=\left\{\boldsymbol{b_{t_1}}, \boldsymbol{b_{t_2}}, ..., \boldsymbol{b_{t_N}}\right\}$ and $\boldsymbol B'=\left\{\boldsymbol{b_{t'_1}}, \boldsymbol{b_{t'_2}}, ..., \boldsymbol{b_{t'_{N'}}}\right\}$ at different scales correspondingly are obtained. Then, two groups of representations $\boldsymbol{\tilde R}=\left\{\boldsymbol{\tilde{r}^1}, \boldsymbol{\tilde{r}^2}, ..., \boldsymbol{\tilde{r}^M}\right\}$ and $\boldsymbol{\tilde R'}=\left\{\boldsymbol{\tilde{r'}^1}, \boldsymbol{\tilde{r'}^2}, ..., \boldsymbol{\tilde{r'}^M}\right\}$ are obtained after the multi-anchor module and time aggregation module with parameter sharing. Finally, we aggregate $\boldsymbol{\tilde R}$ and $\boldsymbol{\tilde R'}$ with self-attention as follow:
\begin{equation}
\label{multiscale}
\boldsymbol{\hat{r}^i}=s^i([\boldsymbol{\tilde{r}^i},\boldsymbol{\tilde{r'}^i}])\odot
\boldsymbol{\tilde{r}^i}+(1-s^i([\boldsymbol{\tilde{r}^i},\boldsymbol{\tilde{r'}^i}]))
\odot\boldsymbol{\tilde{r'}^i},
\end{equation}
where $\odot$ is the Hadamard product, $[\cdot]$ is an operation concatenating vectors along the last dimension, and $s^i(\cdot)$ is the switch function which is implemented as a fully-connected layer followed by a sigmoid function. The switch function controls the fusion between BoI feature sequences of different scales. The output of this module $\boldsymbol{\hat{R}}=[\boldsymbol{\hat{r}^1}, \boldsymbol{\hat{r}^2}, ..., \boldsymbol{\hat{r}^M}]$ is served as the final user representation.

\subsubsection{Contrastive Learning Module}
As mentioned in \ref{contrastive loss}, contrastive loss is also used to learn user representation in this network. In order to obtain a unified vector, a nonlinear projection head $h(\cdot)$ is added, as used in SimCLR \cite{chen2020simple}. The function $h(\cdot)$ is computed as
\begin{equation}
\label{projection}
\begin{split}
\boldsymbol{v}\ \ &= ReLU([\boldsymbol{W_1^{h_1}\hat{r}^1},\boldsymbol{W_2^{h_1}\hat{r}^2},...,\boldsymbol{W^{h_1}_M\hat{r}^M}])\\
h(\boldsymbol{v}) &=\boldsymbol{W^{h_2}} ReLU(\boldsymbol{v}+mix(\boldsymbol{v},\boldsymbol{W^{h_3}}).\\
\end{split}
\end{equation}
The function $mix(\boldsymbol{v},\boldsymbol{W^{h_3}})$ is to allow interaction between different preferences, and is implemented as
\begin{equation}
\label{1}
mix(\boldsymbol{v},\boldsymbol{W^{h_3}})=r_2(\boldsymbol{W^{h_3}}r_1(\boldsymbol{v})),\\
\end{equation}
where $r_1(\cdot)$ reshapes $\boldsymbol{v}$ to a matrix of size $M\times H$, and $r_2(\cdot)$ reshapes $\boldsymbol{W^{h_3}}r_1(\boldsymbol{v})$ to a vector.
 $\boldsymbol{W_i^{h_1}}\in \mathbb{R}^{H\times H}$, $\boldsymbol{W^{h_2}}\in \mathbb{R}^{MH\times MH}$, and $\boldsymbol{W^{h_3}}\in \mathbb{R}^{M\times M}$ are trainable parameters.

Two behavior sequences drawn from the same user are treated as a positive pair, and behavior sequences from other users are negatives. All sequences are continuous sub-sequences randomly sampled from the whole life time $\boldsymbol T$. Finally, the contrastive loss is defined the same as equation (\ref{contrastiveloss}).

\section{Experiments}
In this section, we conduct several experiments to evaluate the performance. The proposed method is compared with several state-of-the-art models on two benchmarks: Amazon dataset \cite{ni2019justifying}  and a real-world industrial dataset. A detailed ablation study is also conducted to validate the effectiveness of each module in LURM.

\subsection{Datasets \& Training Pipeline}
Performance comparisons are conducted on a public dataset and an industrial dataset. Table \ref{dataset_info} shows the statistics of two datasets. The public Amazon dataset contains behavior data like product reviews and metadata like product titles and categories from Amazon. For each user, the reviewed product titles constitute a sequence of review behaviors. The industrial dataset is collected from visit/order logs on a popular e-commerce platform. The whole training process consists of two parts, i.e., training the universal representation model in a self-supervised manner (Stage I) and training an MLP for each specific downstream task (Stage II). The training datasets for stage I are collected as follows. For Amazon, the dataset is constructed by collecting users who have behaviors between 1997-01 and 2017-12.  For industrial dataset, we construct a training dataset by randomly sampling hundreds of millions of users who have behaviors between 2016-06 and 2021-05.
Note that we do not fine-tune the representation model on stage II, and only training simple MLPs for applications are easy, efficient and low-cost.

\begin{table}
\centering 
\caption{Statistics of the datasets. $Tr\left(\cdot \right)$ indicates the truncation threshold.} 
\label{dataset_info}
 \begin{tabular}{cccccc} 
     \toprule
      Dataset & $\left| U\right|$  &  $Max\left(\left| \boldsymbol{S}\right|\right)$ & $Tr\left(\left| \boldsymbol{x}\right| \right)$ & $\left| V\right|$ \\
       \midrule
       Amazon & 43,531,850 & 13,122 & 35 & 103,581\\ 
       Industrial  & 214,317,285 & 1,952,546 & 24 & 178,422&\\ 
       \bottomrule
   \end{tabular}
\end{table}

\begin{table*}
\centering 
\caption{Comparison in terms of AUC(\%)/ACC(\%) on the Amazon dataset. Several category preference identification tasks are evaluated. The best score is bold.} 
\label{comparison_1}
 \begin{tabular}{cccccc} 
     \toprule
      Method & Books of Literature  &  Games of Sports & Outdoor-hunting & Musical Instrument \\
      \midrule
      TextCNN & 73.84/72.83 & 61.00/69.92 & 66.06/72.76 & 68.28/71.81  \\ 
      HAN     & 79.12/77.49 & 66.62/70.01 & 71.19/73.42 & 70.01/72.12 \\
      \hline
      TF-IDF  &78.62/77.21 & 65.69/68.69 & 71.90/72.19 & 69.16/68.62  \\ 
      Doc2Vec &71.21/68.26  & 66.75/66.85 & 71.58/66.36 & 70.29/67.37 \\
     \hline
      PTUM  &  78.66/77.91  & 66.06/68.97  &  70.34/73.01 & 69.50/72.92 \\
      SUMN    &79.57/77.73 & 66.83/70.63 & 72.27/73.44 & 70.61/73.43\\
      \hline
      Ours-LURM & $\boldsymbol{82.55}/\boldsymbol{79.13}$ & $\boldsymbol{68.02}/\boldsymbol{72.43}$ & $\boldsymbol{77.59}/\boldsymbol{75.19}$ & $\boldsymbol{74.04}/\boldsymbol{74.43}$ \\
      \bottomrule
  \end{tabular}
\end{table*}

\subsection{Downstream Tasks}
Two kinds of downstream tasks are used to evaluate our method: category preference identification and user profiling prediction. 

Category preference identification refers to the task of predicting whether users have preferences in the target category, \emph{i.e.}, whether a user would have behaviors on the items of the target category in a future time period. For Amazon, we collect user behaviors from 1997-01 to 2017-12 to form the network inputs, and pass them through LURM to obtain full-life cycle representations (marked as `LURM'). A user is labeled as positive if there exists at least one review log between 2018-01 and 2018-10. There are four categories being included, \emph{i.e.}, `Books of Literature', `Games of Sports', `Outdoor-hunting', and `Musical Instrument'. On the industrial dataset, we collect behaviors of two time periods to infer user representations, aiming to show the effectiveness of full-life cycle modeling, as well as to compare to competitors under the same conditions. Specifically, one deploys behaviors from 2020-06 to 2021-05, marked as `LURM'. The other one uses behaviors from 2021-04 to 2021-05, marked as `LURM(2M)'. The behavior logs between 2021-07 and 2021-08 are used for labeling. We consider three categories including clothing, shoe and coffee. Due to the time and resource limitations, it is impractical to include all categories in the experiments. The selected categories are expected to be representative, covering different scales (namely the number of products in the category), industries, etc.

User profiling prediction aims to identify user properties such as gender and age. We notice that increasing the numbers of behaviors would significantly benefit the performance on user profiling tasks. To show that, we conduct experiments on the industrial dataset, where the number can reach millions. We also collect behaviors of two time periods to infer user representations. One deploys behaviors from 2016-06 to 2021-05, marked as `LURM'. The other one uses behaviors from 2021-04 to 2021-05, marked as `LURM(2M)'. Two specific tasks are involved: (1) user age classification (age is divided into 6 classes) which predicts the age ranges of users, and (2) baby age classification (7 classes). For both tasks, the ground-truth labels are collected from an online questionnaire. 

In the experiments, we randomly select 80\% of the samples for each task to train downstream models and the rest for performance validation. A detailed demonstration of the effectiveness of full-life cycle modeling is given in Section 3.6.

\subsection{Competitors}
We compare our LURM against a rich set of user representation methods, including TF-IDF, Doc2Vec, TextCNN, HAN, PTUM \cite{wu2020ptum} and SUMN \cite{gu2020exploiting}. TF-IDF and Doc2Vec view the whole user behavior sequence as a document. TF-IDF generates a sparse high-dimensional vector, while Doc2Vec learns to represent the document by a dense vector. We also compare LURM with two supervised methods, namely TextCNN and HAN, which are trained particularly for each downstream task.
Specifically, TextCNN adopts convolutional neural networks (CNN) on the embedding sequence of all words appeared in behaviors and uses max-pooling to get the final user representation.
The user representation encoder and the classifier are trained together. 
HAN employs a hierarchical attention network, where two levels of attention operations are used to aggregate words and behavior embeddings respectively. 
Finally, we compare LURM with two newly presented self-supervised methods for user modeling, PTUM and SUMN. 
{PTUM is designed based on BERT and proposes two self-supervision tasks for pre-training. 
The first one is masked behavior prediction, which can model the relatedness between historical behaviors. The second one is next K behavior prediction, which characterizes relatedness between past and future behaviors. 
SUMN proposes a multi-hop aggregation layer to refine user representations and uses behavioral consistency loss to guide the model to extract latent user factors. }

\subsection{Training Details}
\textbf{Item Embedding} As in\cite{gu2020exploiting,wu2020ptum}, the number of words in each item $\boldsymbol{x}$ is truncated. The truncation principle is that 95\% data values can be covered by the chosen threshold. As a result, we set 35 on the Amazon dataset and 24 on the Industrial dataset (see Table \ref{dataset_info}). The window size $\beta$ is set to 5 days\cite{barkan2016item2vec}., and the temperature $\tau$ is set to be 0.1.

\textbf{BoI \& SMEN} We set the latent dimensions of the word/item embeddings and all hidden layers to be 128. The size of the `interest' vocabulary is set to be $10^5$ in order to retain details as many as possible. The number of the anchors in multi-anchor module, namely $M$, is set as 10, and thus the dimension of the final user representation is 1280. We will discuss the performance effects of these two hyper-parameters later. In the multi-scale aggregation module, both the inputs of month-granularity and year-granularity are deployed to capture diverse behavior patterns behind different time periods. We will also compare the performances of using inputs of different granularities in the ablation study section. Moreover, for the user representation model, the loss is optimized by Adam optimizer with a learning rate of 0.001, and a batch size of 256. 

\textbf{Downstream Model} For downstream tasks, a simple MLP classifiers is applied after the derived user representations. The MLP contains only one hidden layer with a dimension of 64. An Adam optimizer with a learning rate of 0.001 is used. The batch size is set as 128. Note that LURM, SUMN and PTUM are not fine-tuned for downstream tasks in our experiments.

\textbf{Competitors} On Amazon dataset, only TF-IDF and Doc2Vec can make use of the entire behavior data to generate representations (same as our method). Meanwhile, the length of input behaviors of other competitors, namely TextCNN, HAN, SUMN and PTUM, is limited to 50 due to memory limitation. On the industrial dataset, all competitors deploy behavior data of two months as inputs, limited by their ability to handle long sequences. For supervised competitors, we use Adam with a learning rate of 0.001 as the optimizer, and the batch size is set as 256. For unsupervised ones, only downstream model needs training. The configurations are set to be the same as in the preceding paragraph.

\begin{table}
\centering 
\caption{Comparison in terms of AUC(\%)/ACC(\%) on Industrial dataset. Two profiling prediction tasks are evaluated. The best score is bold.} 
\label{comparison_2}
 \begin{tabular}{cccccc} 
     \toprule
       Method        & Age         & Baby Age  \\
       \midrule
       TextCNN       & 86.40/61.02 & 72.19/66.01 \\ 
       TF-IDF & 87.73/61.75 & 73.08/67.26  \\ 
       SUMN          & 85.35/60.61 & 74.20/67.63 \\ 
      \hline
       Ours-LURM(2M)   & 88.48/61.99 & 82.71/69.29 \\ 
       Ours-LURM          & $\boldsymbol{96.09}/\boldsymbol{78.43}$ & $\boldsymbol{94.59}/\boldsymbol{84.91}$ \\ 
       \bottomrule
   \end{tabular}
\end{table}

\begin{table}
\centering 
\caption{Comparison in terms of AUC(\%)/ACC(\%) on Industrial dataset. Three category preference identification tasks are evaluated. The best score is bold.} 
\label{comparison_3}
 \begin{tabular}{cccccc} 
     \toprule
       Method      & Clothing    & Shoe        & Coffee\\
       \midrule
       TextCNN     & 76.64/76.13 & 84.33/80.24 & 80.49/79.46  \\ 
       TF-IDF & 77.61/78.01 & 85.04/81.15 & 81.75/80.78  \\ 
       SUMN         & 73.86/74.16 & 83.47/79.63 & 79.89/79.42 \\ 
      \hline
       Ours-LURM(2M)    & 78.37/78.57 & 85.15/81.18 & 81.94/80.78 \\ 
       Ours-LURM         & $\boldsymbol{80.37}/\boldsymbol{79.10}$ & $\boldsymbol{86.30}/\boldsymbol{81.65}$ & $\boldsymbol{83.11}/\boldsymbol{81.18}$ \\ 
       \bottomrule
   \end{tabular}
\end{table}

\subsection{Results}
Table \ref{comparison_1} shows the comparison of category preference identification on the public Amazon dataset. The last row shows the result of our models. It can be seen that LURM consistently outperforms other unsupervised methods, e.g., about 3.23\%/{1.49\%} average improvements than SUMN and 4.41\%/2.09\% average improvements than PUTM in terms of AUC and ACC respectively. Moreover, though our goal is not to beat supervised competitors, LURM still achieves convincing results in comparison with TextCNN and HAN, about 8.26\% and 3.82\% higher average AUC, and about 3.47\% and 2.04\% higher average ACC, respectively. \footnote{BERT-like methods are not included because the number of behaviors (i.e., input tokens) are tremendous, and accordingly the resource costs are too high.} 

Table \ref{comparison_2} and Table \ref{comparison_3} shows the results of two user profiling tasks and three category preference identification tasks on the industrial dataset, respectively. 
Similar observations can be concluded. LURM(2M) achieves consistently better results than any other methods with behavior sequences of the same length. One can also see that further improvements can be achieved by using more behaviors within a longer time period. For example, LURM achieves about 5.66\%/10.16\% average improvements than LURM(2M). Furthermore, we notice that the improvements are more significant on the user profiling tasks. This is probably because the quantity of behaviors related to user profiling is usually larger than that related to category preference prediction. Specifically, incorporating behaviors occurred even years ago can still benefit the performance of user profiling, \emph{e.g.}, age prediction. 

\begin{table*}
\centering 
\caption{Comparison of LURM with different configurations on industrial dataset.}
\label{comparison_scale}
 \begin{tabular}{cccccc} 
     \toprule
      Method              & Age         &  Baby Age   & Clothing    & Shoe        & Coffee\\
       \midrule
       BoI                & 96.48/80.19 & 93.29/83.57 & 80.83/79.33 & 86.69/82.16 & 83.78/81.74 \\
       LURM               & 96.09/78.43 & 94.59/84.91 & 80.37/79.10 & 86.30/81.65 & 83.11/81.18 \\ 
         \hline
       LURM(monthly)      & 95.82/77.54 & 93.19/82.17 & 80.30/79.02 & 86.26/81.61 & 83.08/81.14 \\ 
       LURM(yearly)       & 95.61/77.19 & 92.80/80.93 & 80.24/78.88 & 86.17/81.36 & 82.66/80.78 \\ 
       \hline
       LURM(1anchor-128)  & 93.98/72.78 & 86.13/71.22 & 78.93/78.27 & 85.80/81.11 & 81.98/80.39 \\ 
       LURM(1anchor-1280) & 94.35/73.71 & 89.21/75.35 & 79.38/78.54 & 86.07/81.32 & 82.34/80.59 \\ 
       \bottomrule
   \end{tabular}

\end{table*}

\begin{figure*}
\begin{center}
\centerline{\includegraphics[width=\textwidth]{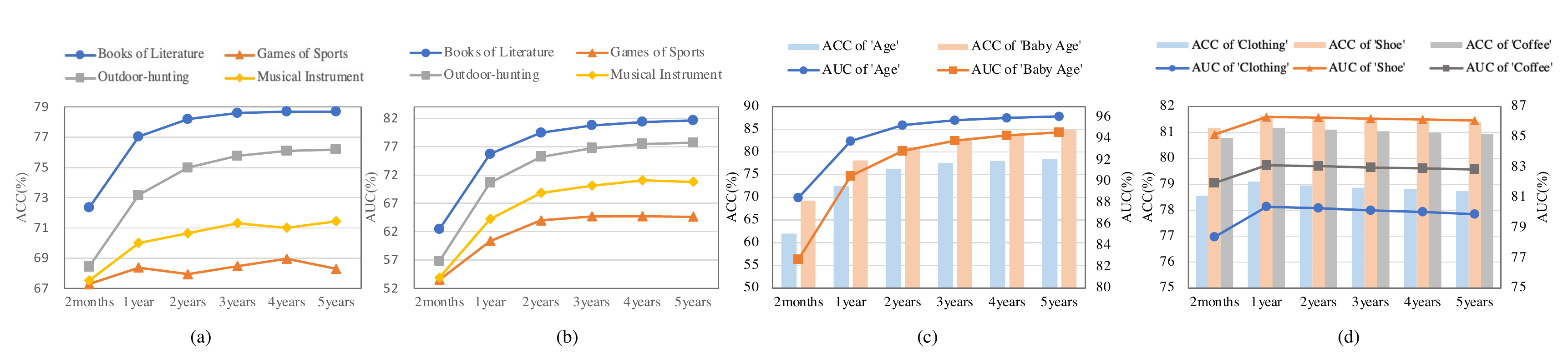}}
\caption{Comparison of LURM with  inputs of different lengths on several downstream tasks. (a-b) show the performance on the Amazon Dataset, and (c-d) show the results on the Industrial Dataset}
\label{comparison_length}
\end{center}
\end{figure*}

\subsection{Ablation Studies \& Discussion} 
Fig. \ref{comparison_length} shows the performances of LURM on several downstream tasks of two datasets when using inputs of different lengths.
It can be seen from Fig. \ref{comparison_length}(c) that the richer the behavior, the better the performance on user profiling. Compared to using only behaviors of two months, an average improvement of 9.75\%/16.03\% can be achieved when using behaviors of 5 years on these tasks. While on category preference prediction tasks of both datasets, there are noticeable improvements in the early stage, and then the improvements gradually decrease. The best performance is obtained when using behaviors with a length of one or two years. It may because that the information gain is limited when mining patterns from extremely long behavior sequences for these tasks.

Table \ref{comparison_scale} shows the results of LURM with different configurations on industrial dataset. The first row shows the results of using high-dimensional BoI features as user representations in downstream tasks. It can be seen that the difference between BoI and LURM is tiny, which proves that SMEN can achieve almost lossless dimensionality reduction.

We also verify the necessity of multi-scale aggregation module. The third and fourth rows of Table \ref{comparison_scale} show the results of LURM with input at the monthly/yearly granularity only, respectively. It can be seen that using multi-scale aggregation module can achieve significant improvements on user profiling prediction tasks, while the benefits are relatively marginal disappear on category preference prediction tasks. The reason is similar, refer to our explanations of the results in Fig. \ref{comparison_length}. 

The last two rows of Table \ref{comparison_scale} show the results of LURM with 1 anchor, where the dimensions are set as 128 and 1280 respectively. Taking BoI as a benchmark, it can be seen that LURM achieves almost lossless dimensionality reduction compared to LURM(1anchor-128) and LURM(1anchor-1280), which verifies the effectiveness of the multi-anchor module.

\begin{figure*}
\begin{center}
\centerline{\includegraphics[width=\textwidth]{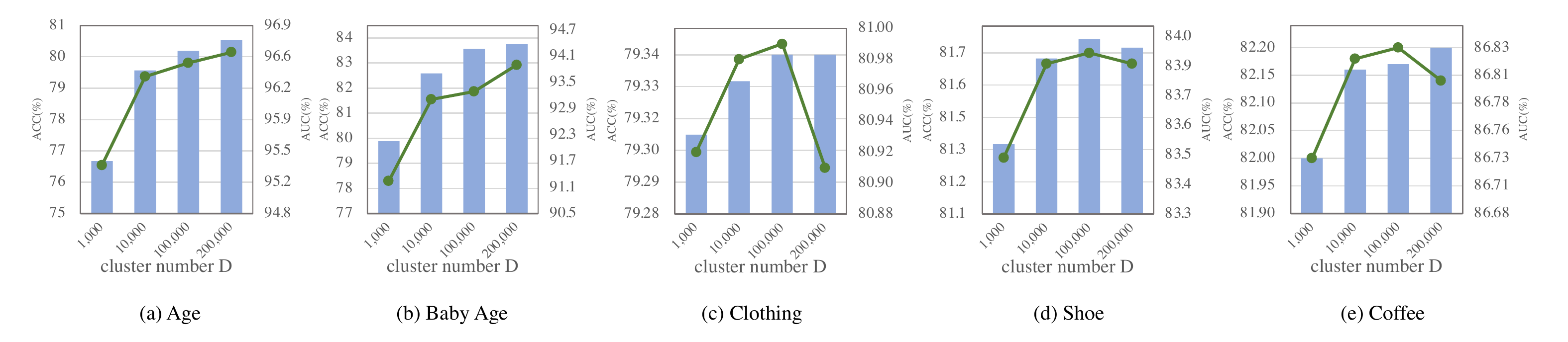}}
\caption{Performance of different cluster number D in downstream tasks. The green line denotes the AUC score, and the blue bars represent the ACC }
\label{cluster_num}
\end{center}
\end{figure*}

The performance effects of the cluster number D are shown in Fig. \ref{cluster_num}. The experiments are conducted on the industrial dataset. In general, one can see that the performance improves if we increase D from 1000 to 100000, while the improvements fade away when D reaches 200000.

We also explore the performance effect of the cluster assignment strategy of items. To this end, a soft assignment strategy is tested, \emph{i.e.}, assigning a item to multiple clusters according to cosine distance. According to our observations, the performance difference of these two strategies is less than 1\%. Since soft assignment would bring extra storage and calculation costs, hard assignment is adopted in this paper.

\subsection{Visualization of Representation}
\begin{figure}[ht]
\begin{center}
\centerline{\includegraphics[width=0.5\textwidth]{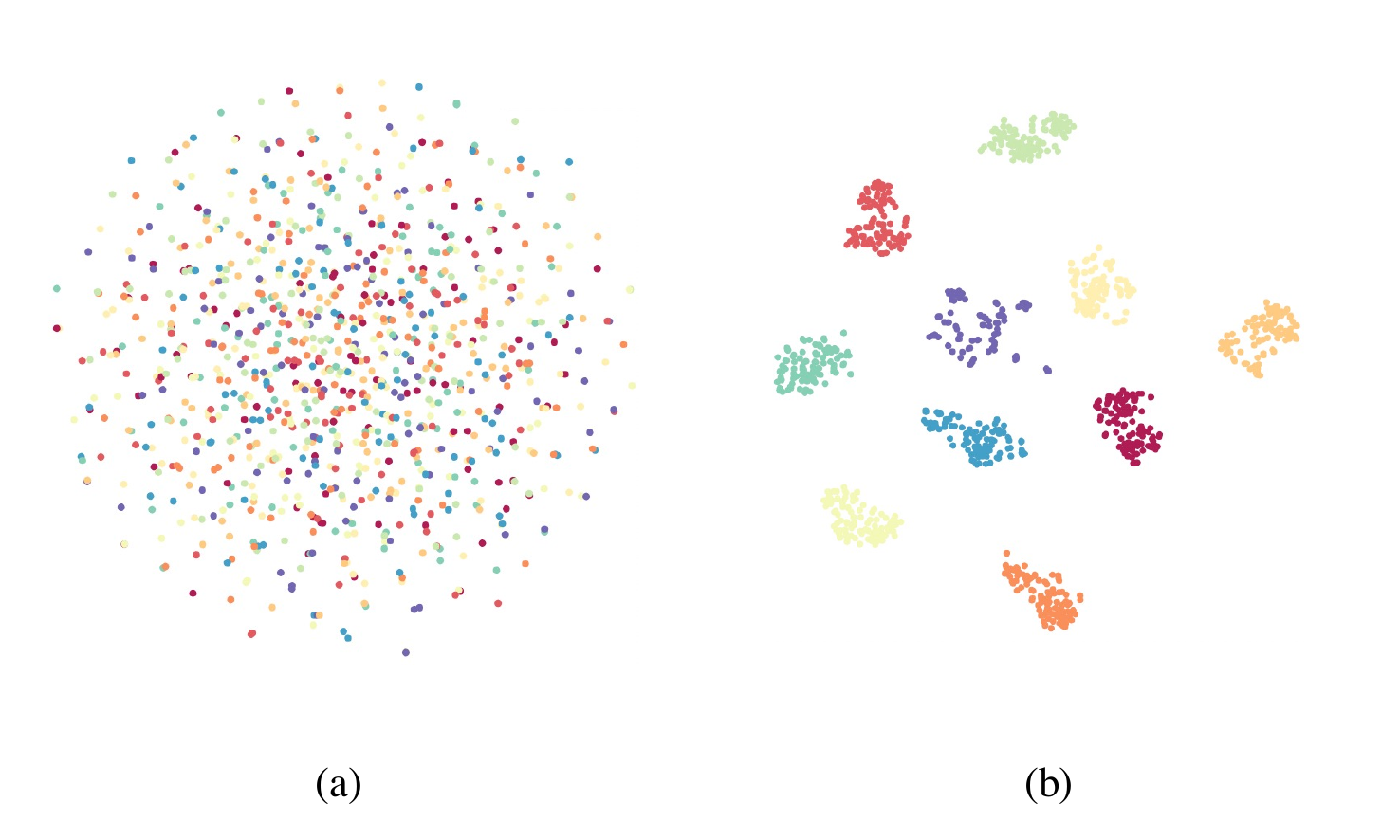}}
\caption{t-SNE visualization of user representations. The points of the same color indicate that they come from the same anchor. (a) representations after model initialization, (b) representations learning by multi-anchors after training.}
\label{t-sne}
\end{center}
\end{figure}

The representations learned on the industrial dataset are intuitively illustrated in Fig. \ref{t-sne}. We visualize the outputs of the multi-anchor module for clarity, which are mapped into the 2-dimensional space with t-SNE\cite{2008Visualizing} (including representations generated by 10 anchors from 100 randomly selected users). Different colors correspond to difference anchors. Visually, this illustration demonstrates that LURM can gradually mine and characterize different aspects of user interests, showing great diversity.

\subsection{Comparison to Concurrent Methods}
This work has been applied widely in Alibaba. During the process, more experiments are conducted on our own dataset, including comparisons with concurrent methods \cite{2021Scaling,2021Interest}. 

Though we cannot release the datasets and specific results, the conclusions and observations can be shared. Generally, LURM have similar performance with ICL\cite{2021Interest}, but LURM (long) (namely incorporating more behaviors) achieves about 2\%-19\% higher AUC than ICL. We also have compared our method to a CLIP-like model, which is similar to \cite{2021Scaling}. The observations are similar, due to the capacity of large model, the CLIP-like model performs slightly better than LURM, but LURM (long) shows superior performance, outperforming CLIP-like model by about 1\%-3\%.

\section{Related Works}
\subsection{Universal User Modeling}
Compared with task-specific user modeling that requires more resources, universal user representations are preferred to serve different downstream tasks. In recent years, some works dedicated to learning universal user representations have been proposed \cite{robertson2004understanding,ding2017multi,ni2018perceive,andrews2019learning,gu2020exploiting,wu2020ptum,wu2022userbert,2021Scaling,2021Interest}.
Ni \textit{et al.} \cite{ni2018perceive} proposed a representation learning method based on multi-task learning, which enabled the network to generalize universal user representations. Extensive experiments showed the generality and transferability of the user representation. However, the effectiveness of this method may still suffer due to the selection of tasks and the need of labels. To release the burden of labeling, Andrews \textit{et al.} \cite{andrews2019learning} proposed a novel procedure to learn user embedding by using metric learning. They learned a mapping from short episodes of user behaviors to a vector space in which the distance between points captures the similarity of the corresponding users invariant features. Gu \textit{et al.} \cite{gu2020exploiting} proposed a network named self-supervised user modeling network (SUMN) to encode user behavior data into universal representation. They introduced a behavior consistency loss, which guided the model to fully identify and preserve valuable user information under a self-supervised learning framework. Wu \textit{et al.} \cite{wu2020ptum} proposed pre-trained user models (PTUM), which can learn universal user models based on two self-supervision tasks for pre-training. The first one was masked behavior prediction, which can model the relatedness between historical behaviors. The second one was next K behavior prediction, which can model the relatedness between past and future behaviors. 
{With the help of self-supervised pretext tasks, these methods can obtain universal user representation that can serve different downstream tasks.} 
Unfortunately, these methods can only process user behavior sequences with a length of thousands (e.g., the largest length of behavior sequences used in \cite{2021Scaling} is 2048, and the length is truncated to 320 in \cite{2021Interest}), and cannot leverage the rich information brought by full-life cycle user behaviors. 

\subsection{Full-life Cycle User Modeling}
Previous works {of task-specific user modeling} have shown that considering long-term historical behavior sequences for user modeling can significantly improve the performance of different tasks \cite{ren2019lifelong,pi2019practice,pi2020search,cao2022sampling}. Ren \textit{et al.} \cite{ren2019lifelong} proposed a hierarchical periodic memory network for full-life cycle sequential modeling. They built a personalized memorization for each user, which remembers both intrinsic user tastes and multi-facet user interests with the learned while compressed memory. Pi \textit{et al.} \cite{pi2019practice} decoupled the user modeling from the whole CTR prediction system to tackle the challenge of the storage cost and the system latency. Specifically, they proposed a user interest center module for real-time inference and a memory-based network that can be implemented incrementally. Pi \textit{et al.} also designed a search-based interest model (SIM) with a cascaded two-stage search paradigm to capture the diverse long-term interest with target item. Unfortunately, the length of the user behavior sequence that these models can handle is still limited. Moreover, these models are all trained on specific tasks(e.g., CTR), which limits the generalization ability.

To the best of our knowledge, we are the pioneer to make full-life cycle modeling possible for learning general-purpose user representation. LURM allows for encoding even millions of historical behaviors to improve the quality of user representations.

\section{Conclusion}
In this work, a novel framework named LURM is proposed to model full-life cycle user behaviors with any length. 
With the ability to model full-life cycle user behaviors, 
our method shows promising results on different downstream tasks and datasets. Although our method has made some progress, there is still space for improvement. In the future research work, we will consider more types of tasks; input data in different modalities, such as images, video, and audio; more dedicated network architecture and so on.


\bibliographystyle{ACM-Reference-Format}
\balance
\bibliography{references}










\end{document}